\documentclass{article}
\usepackage{spconf,amsmath,graphicx,arydshln}
\usepackage{caption, float}

\title{Decoding visemes: improving machine lip-reading}
\name{Helen L. Bear and Richard Harvey}

\address{School of Computing Sciences, University of East Anglia, Norwich, NR4 7TJ, United Kingdom}
\begin{document}
%
\maketitle
\begin{abstract}
To undertake machine lip-reading, we try to recognise speech from a visual signal. Current work often uses viseme classification supported by language models with varying degrees of success. A few recent works suggest phoneme classification, in the right circumstances, can outperform viseme classification. In this work we present a novel two-pass method of training phoneme classifiers which uses previously trained visemes in the first pass. With our new training algorithm, we show classification performance which significantly improves on previous lip-reading results.  

\end{abstract}
\section{Introduction}
\label{sec:intro}
In machine lip-reading, the classification of an utterance from a visual-only signal, there are many obstacles to overcome. Some, such as pose \cite{4218129, 6298439}, motion \cite{927467,ong2011robust} and resolution \cite{bear2014resolution} have been studied and measured, including the selection of a phoneme-to-viseme mapping \cite{cappelletta2012phoneme, bear2014phoneme}. However, visemes are not precisely defined. Many working definitions have been offered such as; ``A set of phonemes that have identical appearance on the lips'' \cite{bear2014phoneme} or ``A visual equivalent of a phoneme'' \cite{bear2014some}. However, there are challenges with using viseme labelled classifiers including: the homophone effect, not enough training data per class, and the consequential lack of differentiation between classes when there are too many visemes within a set. More recently, there is evidence that viseme labels may not be needed at all because with enough data, classifiers based on phoneme labels can outperform viseme classification \cite{howellPhD, hazen2006visual}. As phonemes are well studied, this idea is attractive. However, others have tested small numbers of visual units: visemes and found they also give acceptable results \cite{7074217, bozkurt2007comparison}. It would be very helpful to be able to systematically vary the number visual units and hence devise optimal strategies for learning. 

%

The rest of this paper is as follows; a summary of the analysis into the effect of varying the quantity of visemes in a set on lip-reading performance presented in \cite{bear2015findingphonemes} is followed by a short test on unit selection effects between classifier and its supporting network, the results of these are used to introduce the hypothesis for applying weak learning during classifier training. A full description of the experimental setup to test the hypothesis is included before analysis of results and conclusions. 

\section{Background}

A systematic study into varying the number of visemes was conducted in \cite{bear2015findingphonemes} which generated viseme sets of varying size. HTK \cite{young2006htk} was used to build Hidden Markov Model (HMM) classifiers for every viseme in each set. We initialised a set of HMMs (\texttt{HCompV}), that were trained (and retrained) using \texttt{HERest} during which there were options to tie any required model states together (e.g. for short pause models) (\texttt{HHed}) or to force align the HMMs to a time-aligned ground truth (\texttt{HVite}) before producing a classification output. The output of classification was supported by a word bigram model created with \texttt{HBuild} and \texttt{HLStats}. Finally, this classification output was compared to the ground truth to measure its efficacy (\texttt{HResults}) which we measured using Correctness, $C$. 

\begin{equation}
	C = \displaystyle \frac{N-D-S}{N},\quad
        	\label{eq1}
\end{equation}
where $S$ is the number of substitution errors, $D$ is the number of deletion errors, $I$ is the number of insertion errors and $N$ the total number of labels in the reference transcriptions \cite{young2006htk}.

\begin{figure}[H]
\begin{minipage}[b]{1.0\linewidth}
  \centering
  \centerline{\includegraphics[width=8.5cm]{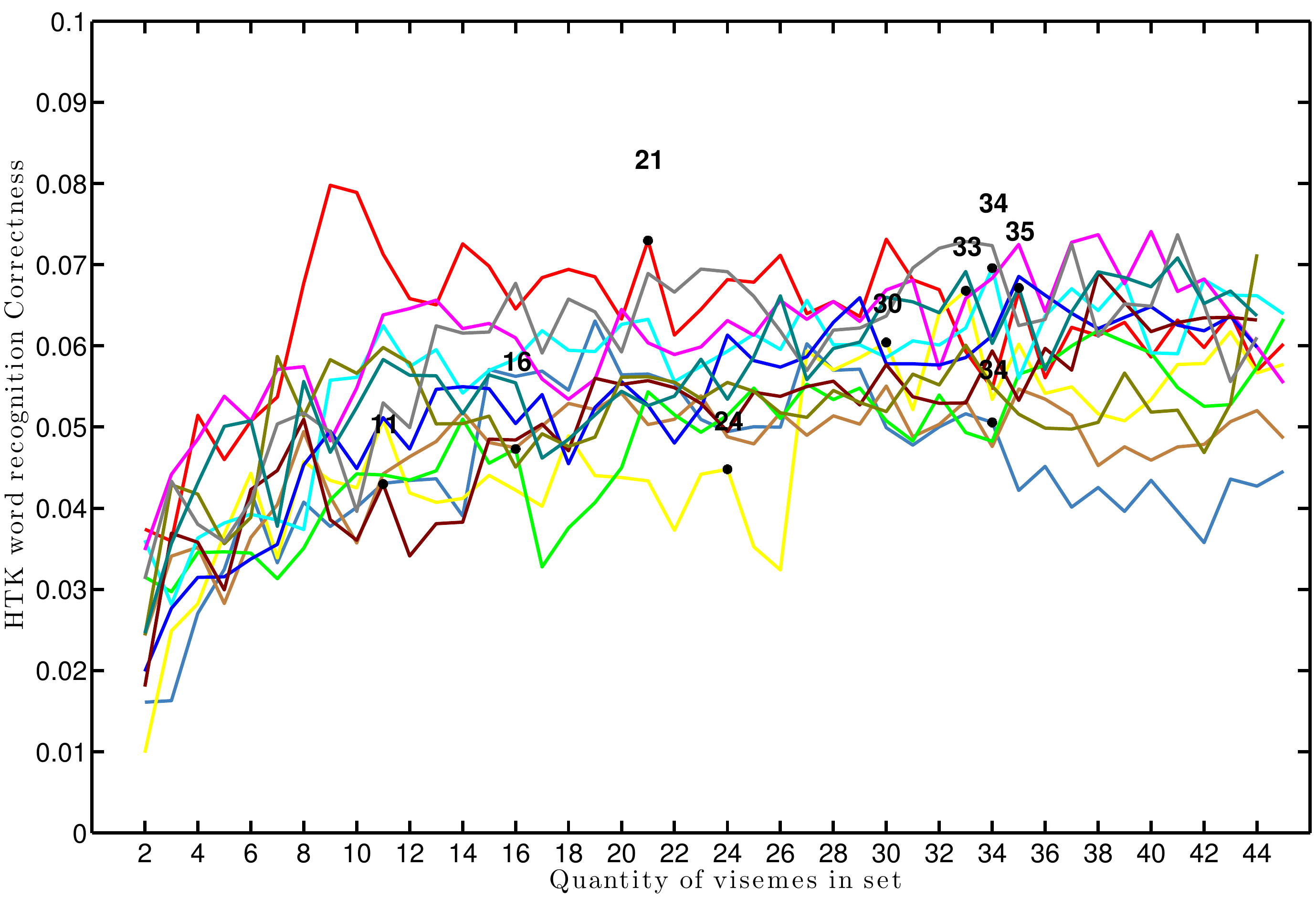}}
  \centerline{}\medskip
\end{minipage}
 \vspace{-2.5em}
\caption{Viseme correctness as the quantity of visemes changes in a set of classifiers for 12 LiLiR speakers. Results from \cite{bear2015findingphonemes}.}
\label{fig:biggraph}
\end{figure}

Figure~\ref{fig:biggraph} shows our previous results \cite{bear2015findingphonemes}, derived using the algorithm described in \cite{bear2014phoneme}. The algorithm works by merging visemes. For example, a label set with 44 visemes has been obtained from the label set of 45 visemes. At each merging stage we measure the difference in correctness compared to the previous set. Significant differences in Figure~\ref{fig:biggraph} are shown with black dots where the number represents the size of the significant set. 


In Figure~\ref{fig:biggraph} the performance of classifiers with few visemes is poor due to the large number of homophones. An example of a homophone in the data are the words ``port'' and ``bass''. Using Speaker 1's 10-viseme P2V map these both become `$/v5/$ $/v9/$ $/v7/$' i.e. a single identifier for identifying two distinct words. Thus distinguishing between ``port'' and ``bass'' is impossible. Large numbers of visemes do not appear to further improve the correctness, probably because, as has been observed before, many phonemes look similar on the lips \cite{fisher1986darpa}. Looking at Figure~\ref{fig:biggraph} there appears to be a sweet spot where optimality might be found between visemes set sizes from $11$ to $36$. 

\section{Data}
For comparable experiments, we select the same 12 speakers from the dataset \cite{improveVis} presented in \cite{bear2015findingphonemes}. For the seven male and five female speakers, each utters 200 sentences from \cite{fisher1986darpa}. Individual speakers were tracked using Active Appearance Models (AAMs) \cite{Matthews_Baker_2004} and the extracted features consist of concatenated shape and appearance information representing only the mouth area of the face. 

\section{Method}

\begin{figure}[htb]
\begin{minipage}[b]{1.0\linewidth}
  \centering
  \centerline{\includegraphics[width=8.8cm]{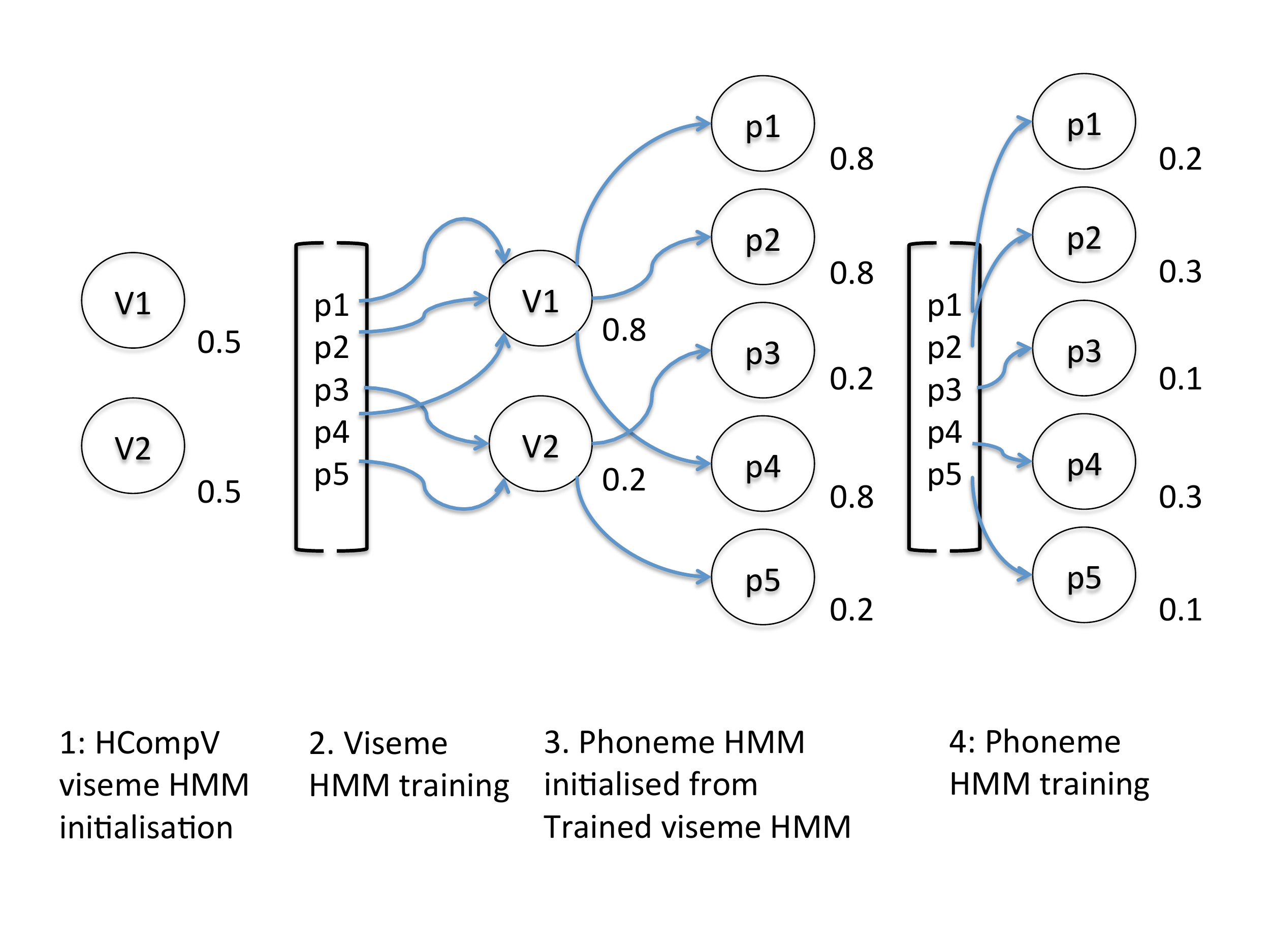}}
  \centerline{}\medskip
\end{minipage}
 \vspace{-5em}
\caption{Weak learning of visemes to initialise phoneme labelled classifiers.}
\label{fig:wlt_process}
\end{figure}

In previous work, we essentially examined two different algorithms. In the first, the data were labelled with phonemes, we use \texttt{HCompV} to initialise the phoneme classifiers, and 11 repetitions of \texttt{HERest} to train the classifiers. This system had the advantage that the output was a sequence of phonemes, but the disadvantage that phoneme models are hard to train. The alternative was to use a smaller number of visemes. The data were labelled with the visemes, and we learned the viseme classifiers in the same way, \texttt{HCompV} followed by \texttt{HERest}. Our new method is a hybrid. We initially learn the visemes, these trained visemes then become the starting point phoneme classifiers (we know the mapping from the visemes to the phonemes for all sets of visemes). We now train the the phoneme models via repeated applications of \texttt{HERest}, thus we have obtained phoneme models but with a new initialisation based upon what was learned for the visemes. This process is illustrated in Figure~\ref{fig:wlt_process}. In this example $p1$, $p2$ and $p4$ are associated with $v1$, so are initialised as replicas of HMM $v1$. Likewise $p3$ and $p5$ are initialised as replicas of $v2$. We now retrain the phoneme models using the same training data.



In full; we initialise \textit{viseme} HMMs with \texttt{HCompV}. Our prototype HMM is based upon a Gaussian mixture of five components and three states \cite{982900}. These are re-estimated 11 times over with \texttt{HERest}, including both short pause model state tying (between re-estimates 3 \& 4 with \texttt{HHed}), and forced alignment between re-estimates 7 \& 8 with \texttt{HVite}. This is steps 1 \& 2 in Figure~\ref{fig:wlt_process}. But before classification, these viseme HMM definitions are used as initialised definitions for phoneme labelled HMMs (Figure~\ref{fig:wlt_process} step 3). The respective viseme HMM definition is used for all the phonemes in its relative phoneme-to-viseme mapping. These phoneme HMMs are retrained and used for classification. This amendment to training is analogous with weak learning. We complete classification twice. First with a phoneme bigram network, second with a word bigram network. For both we apply a grammar scale factor of $1.0$ and a transition penalty of $0.5$ (based on \cite{howellPhD}) with \texttt{HVite}. This is implemented using 10-fold cross-validation with replacement \cite{efron1983leisurely}. %

The advantage of our new training approach is that the phoneme classifiers have seen only positive cases therefore have good mode matching, the disadvantage is they are not exposed to negative cases to the same degree as the visemes. 

\subsection {Language network units}
\label{sec:ln}

The systems under study in this paper have two components. The first component, the classifier takes the raw data and attempts to estimate a probable string of units. The second component, the language model, modifies that string on the basis of knowledge of how the units are co-located in the training data. In practice of course, these two components work together and there is no intermediate uncorrected string. 

Here we are considering the problem of what the classification unit should be: a viseme? A phoneme? Or a word? But we also must consider how the language model should work. Should we use $n$-grams of phonemes? Visemes? Or words? The further confusion is the unit on which we measure correctness. It is possible, for example, to build a word classifier followed by a bigram word network measured in terms of its viseme correctness. Such a system would be bizarre but is none-the-less possible. Table~\ref{tab:sn_tests} shows some of the more sensible possibilities. The first row of Table~\ref{tab:sn_tests} is a viseme classifier followed by a viseme bigram network with a viseme correctness of $0.0231$. In Table~\ref{tab:sn_tests} correctness is always measured by the units of the classifier. The dashed lines group different correctness units. The top group show viseme correctness which can be compared against each other, the second group show phoneme correctness and the bottom, word correctness. 

\begin{figure}[!h]
\begin{minipage}[b]{1.0\linewidth}
  \centering
  \centerline{\includegraphics[width=8.5cm]{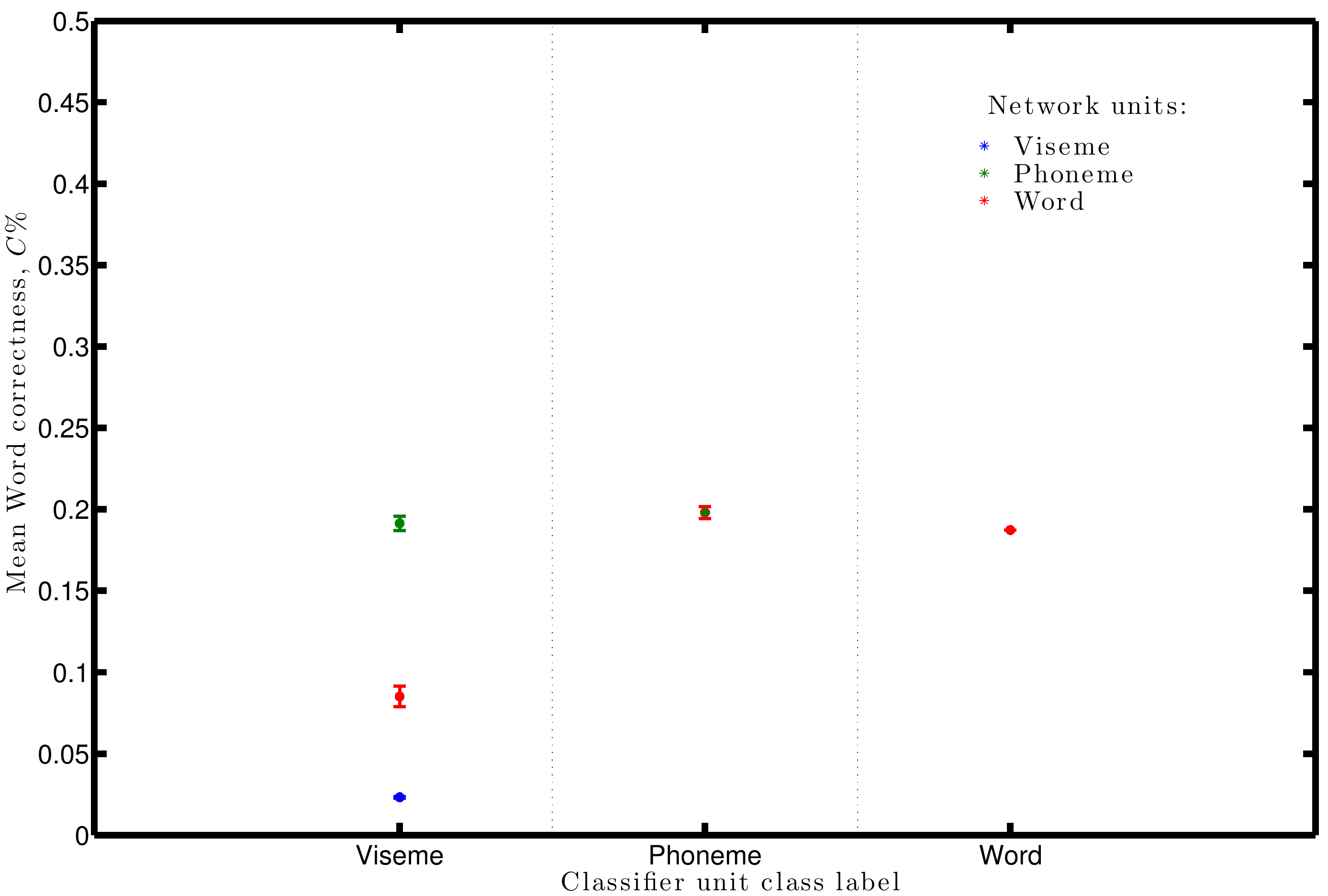}}
 \vspace{-1.5em}
  \centerline{}\medskip
\end{minipage}
\caption{Effects of support network unit choice with varying HMM classifier units measured in all speaker mean correctness, $C$.}
\vspace{-1.5em}
\label{fig:sn_effects}
\end{figure}

In our data we have a large vocabulary (approximately 1000 words), so we eliminate word level classifiers as impractical. This leaves us with viseme classifiers for which the viseme word network is the worst performing so we do not consider this option either. For convenience the same data are plotted in Figure~\ref{fig:sn_effects} with error bars of one standard error. 

\begin{table}[h]
\centering
\caption{Unit selection pairs for HMMs \& language networks.} 
\begin{tabular}{|l|l|l|}
\hline
Classifier units & Network units & Classifier unit, $C$ \\
\hline \hline
Viseme & Viseme & 0.0231 \\
Viseme & Phoneme & 0.1914\\
Viseme & Word & 0.0851 \\
\hdashline
Phoneme & Phoneme & 0.1980\\
Phoneme & Word & 0.1980 \\
\hdashline
Word & Word & 0.1874 \\
\hline
\end{tabular}
\label{tab:sn_tests}
\end{table}

\vfill\pagebreak
\section{Results}

\begin{figure}[h]
\begin{minipage}[b]{1.0\linewidth}
  \centering
	\includegraphics[width=8.5cm]{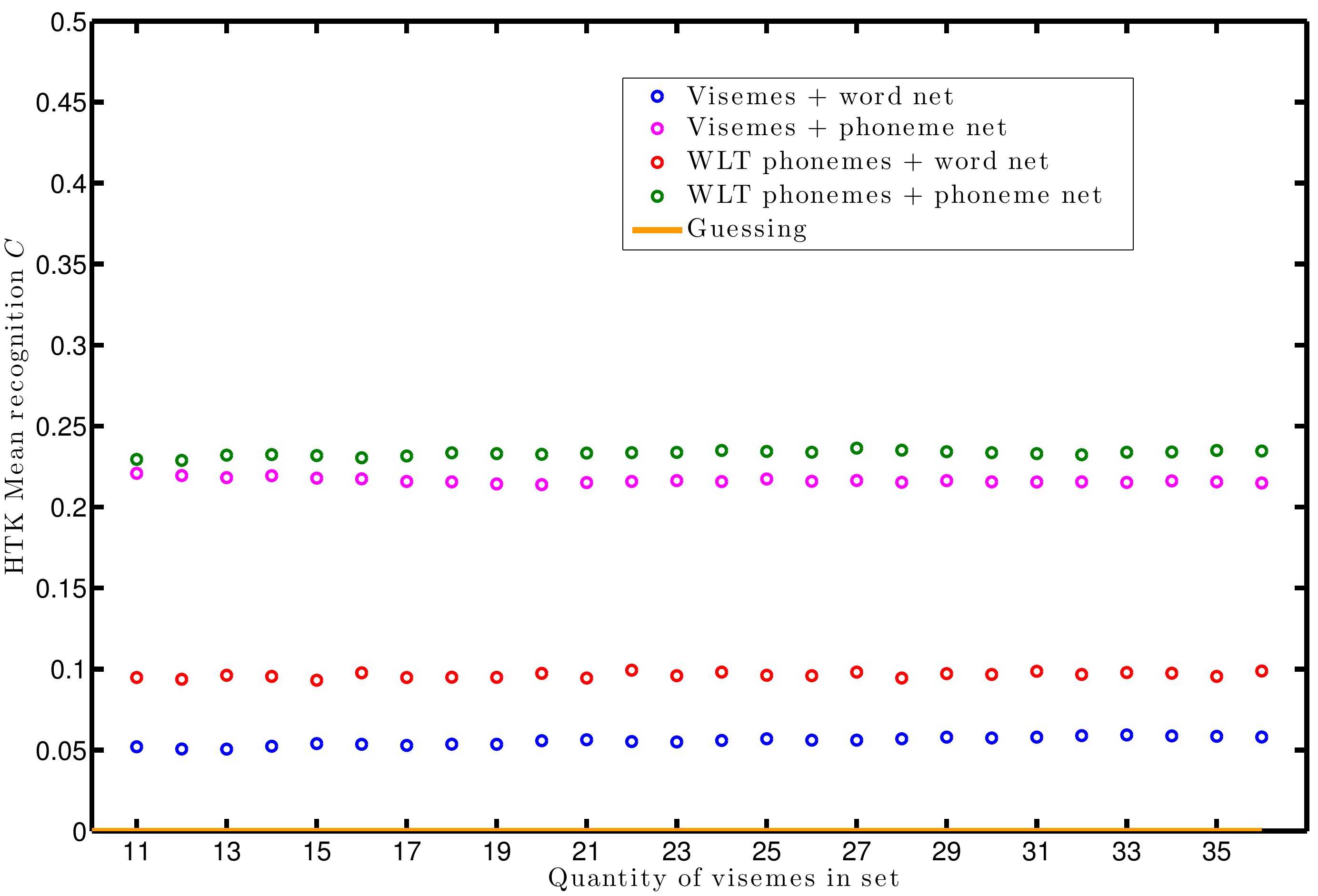}
\end{minipage}	
\caption{HTK Correctness $C$ for both types of classifier with either a phoneme or a word language model averaged over all 12 speakers.} 
\label{fig:res_all}
\end{figure}

Figure~\ref{fig:res_all} shows the mean speaker-dependent correctness. We examine two configurations, one is phoneme classification where we measure phoneme correctness. These are the top two data series in Figure~\ref{fig:res_all} (in green and pink), and the other is word classification where we measure word correctness. These are the lower two data series in Figure~\ref{fig:res_all} in blue and red. Word correctness guessing is duplicated from \cite{bear2015findingphonemes} and is plotted in orange.
 
In the top two series, both have bigram phoneme networks, the lower of these two series uses a viseme classifier as in \cite{bear2015findingphonemes}, and the upper our new phonemes denoted WLT. The lower pair of series use bigram word networks and again show the difference between visemes and our new method of training phoneme classifiers. 
 
The situation in Figure~\ref{fig:res_all} is summarised in Table~\ref{tab:meanstats}. For hard to classify speakers, the new model training method gives a significant improvement.



\begin{table}[h]
\centering
\caption{Minimum, maximum, and range of mean correctness measured over all speakers for the various methods. Top of table shows word correctness, bottom of table phoneme correctness.}
\begin{tabular}{|l|r|r|r|}
\hline
 	& Min & Max & Range \\
\hline \hline
WLT phonemes + phoneme net & 0.2253 & 0.2367 & 0.0114 \\ 
Visemes + phoneme net & 0.2036 & 0.2214 & 0.0179 \\
Effect of WLT & 0.0217 & 0.0153 & -- \\
\hline \hline
WLT phonemes + word net & 0.0905 & 0.0995 & 0.0090 \\
Visemes + word net & 0.0274 & 0.0601 & 0.0327 \\
Effect of WLT & 0.0631 & 0.0394 & -- \\
\hline
\end{tabular}
\label{tab:meanstats}
\end{table}

\begin{figure}[p]
\begin{minipage}[b]{1.0\linewidth}
  \centering
  \centerline{\includegraphics[width=8.5cm]{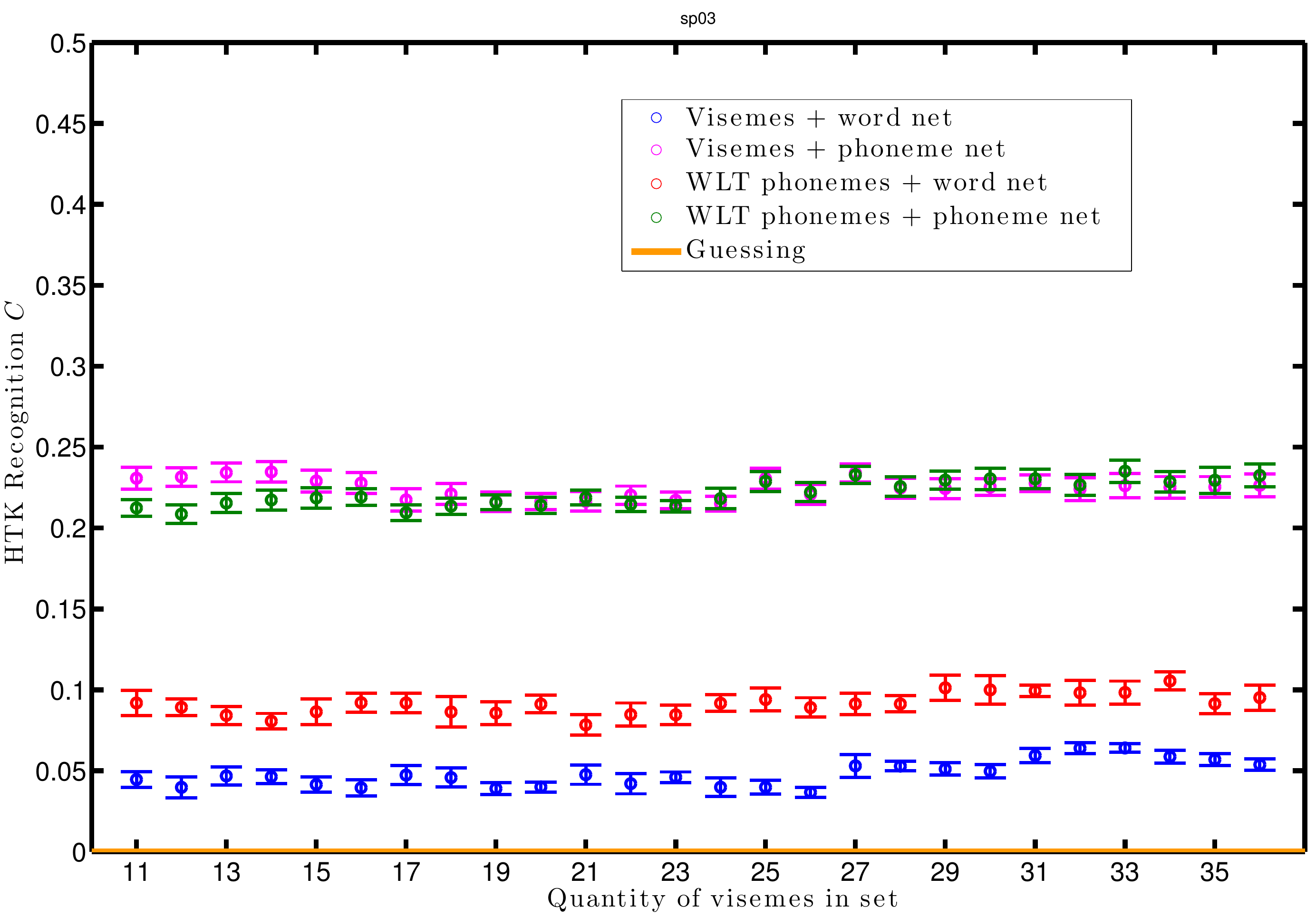}}
  \centerline{(a) Speaker 3}\medskip
  \centerline{\includegraphics[width=8.5cm]{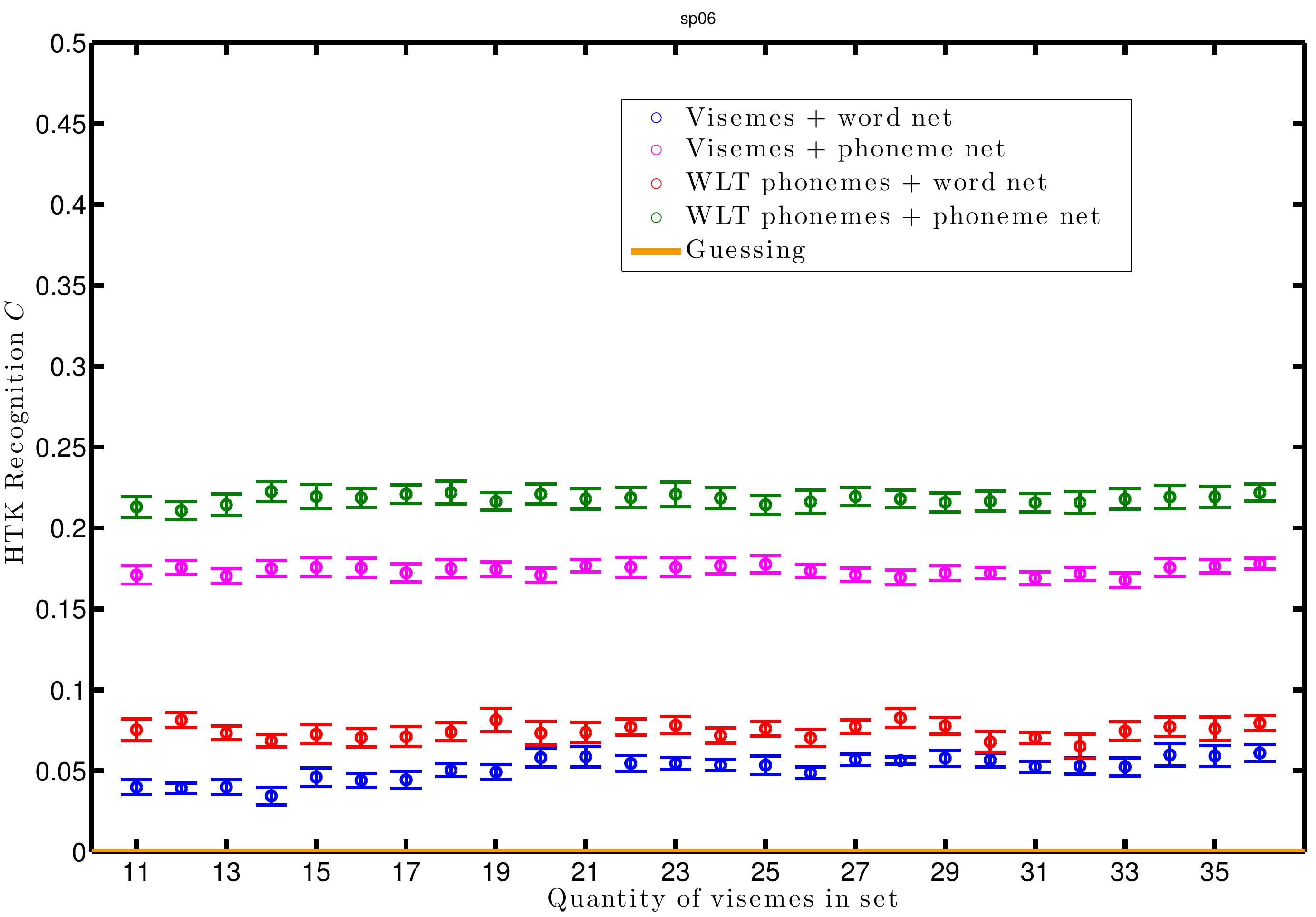}}
  \centerline{(b) Speaker  6}\medskip
  \centerline{\includegraphics[width=8.5cm]{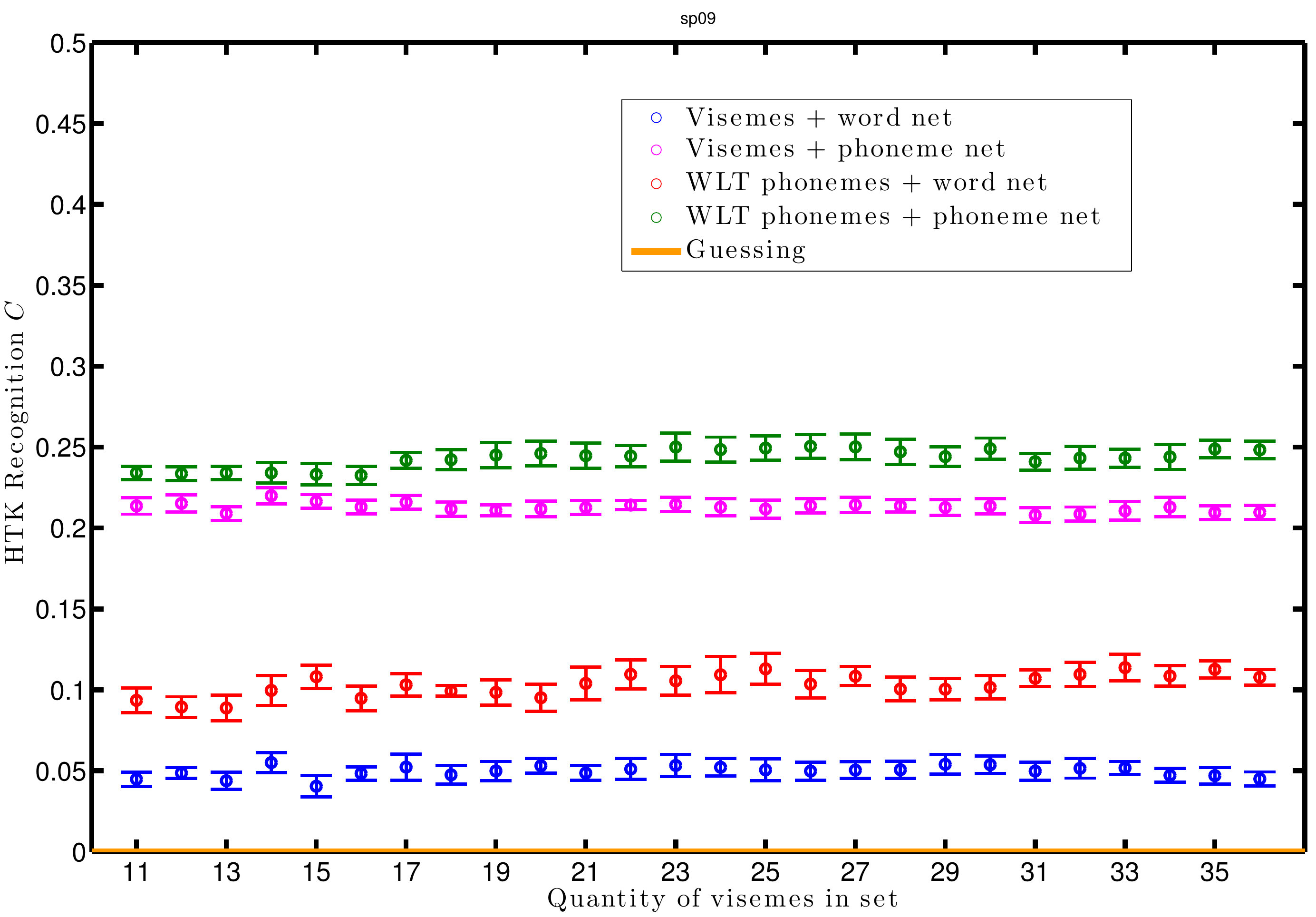}}
  \centerline{(c) Speaker  9}\medskip
\end{minipage}
\caption{HTK Correctness $C$ for a variety of classifiers with either phoneme or word language models for three speakers.}
\label{fig:res}
\end{figure}

\vfill\pagebreak
Figures~\ref{fig:res}a, b \& c show example performances for three speakers. Whilst not monotonic, these graphs are much smoother than the speaker-dependent graphs shown in \cite{bear2015findingphonemes}. Which is encouraging because it implies that our new algorithm is optimising its learning for each speaker-dependent phoneme-to-viseme mapping. 

Figure~\ref{fig:res} shows that, for certain numbers of visemes, and for certain speakers, the weak learning method gives improvement. However, with the right number of visemes for a particular speaker, the new method will always give a significant improvement. 

Looking at Figure~\ref{fig:res} there appeard to be a few regions where the new training method gives only marginal improvement. Not all speakers have these regions. We think the presence of these regions is associated with speakers that have more co-articulation than others. If this is true, then the phonemes are blurred together, the learning is more difficult and performance declines. We do not have enough speakers to make this anything other than speculation at this stage. Our own observation is that young people have more co-articulation than old people, but this is something for further investigation.


%

\section{Conclusions}

The choice of visual units in lip-reading has caused some debate. Some workers use visemes as adduced by for example Fisher \cite{fisher1968confusions} (in which visemes are a theoretical construct representing phonemes should look identical on the lips \cite{hazen2006visual}). Others have noted that lip-reading using phonemes gives superior performance to visemes such as in \cite{howellPhD}. 

Here, we supply further evidence to the more nuanced hypothesis first presented in \cite{bear2015findingphonemes}, which is that there are intermediary units, which for convenience we call visemes, that can provide superior performances provided they are derived by an analysis of the data. A good number of visemes in a set is higher than previously thought.   

In this paper we have presented a novel learning algorithm which shows improved performance for these new data-driven visemes by using them as an intermediate step in training phoneme classifiers. The essence of our method is to re-train the viseme models in a fashion similar to weak learning. This two-pass approach on the same training data has improved the training of phoneme labelled classifiers and increased the classification performance. 




\bibliographystyle{IEEEbib}
\bibliography{refs}

\end{document}